\documentclass[letterpaper]{article} 
\usepackage{aaai23}  
\usepackage{times}  
\usepackage{helvet}  
\usepackage{courier}  
\usepackage[hyphens]{url}  
\usepackage{graphicx} 
\usepackage{natbib}  
\usepackage{caption} 
\usepackage{algorithm}
\usepackage{algorithmic}
\usepackage[english]{babel}
\usepackage{amsmath}
\usepackage{amssymb}
\usepackage{mathtools}
\usepackage{amsthm}
\usepackage[textsize=tiny]{todonotes}
\usepackage{multirow}
\usepackage{microtype}
\usepackage{subfigure}
\usepackage{booktabs}

\usepackage{xcolor}

\usepackage{newfloat}
\usepackage{listings}
\DeclareCaptionStyle{ruled}{labelfont=normalfont,labelsep=colon,strut=off} 
\floatstyle{ruled}
\newfloat{listing}{tb}{lst}{}
\floatname{listing}{Listing}
\title{THMA: Tencent HD Map AI System for Creating HD Map Annotations}
\author{
    Kun Tang\equalcontrib \textsuperscript{\rm 1},
    Xu Cao\equalcontrib \textsuperscript{\rm 1,2}, 
    Zhipeng Cao \equalcontrib \textsuperscript{\rm 1},
    Tong Zhou\textsuperscript{\rm 1},
    Erlong Li\textsuperscript{\rm 1},
    Ao Liu\textsuperscript{\rm 1}, \\
    Shengtao Zou\textsuperscript{\rm 1},
    Chang Liu\textsuperscript{\rm 1},
    Shuqi Mei\textsuperscript{\rm 1},
    Elena Sizikova \correspondingauthors \textsuperscript{\rm 2},
    Chao Zheng \correspondingauthors \textsuperscript{\rm 1}
}
\affiliations{
    \textsuperscript{\rm 1}T Lab, Tencent, Beijing, China\\

    \textsuperscript{\rm 2}Center of Data Science, New York University, New York, United States \\
    es5223@nyu.edu, chrisczheng@tencent.com
}

\usepackage{bibentry}

\begin{document}

\maketitle

\begin{abstract}


Nowadays, autonomous vehicle technology is becoming more and more mature. Critical to progress and safety, high-definition (HD) maps, a type of centimeter-level map collected using a laser sensor, provide accurate descriptions of the surrounding environment. The key challenge of HD map production is efficient, high-quality collection and annotation of large-volume datasets. Due to the demand for high quality, HD map production requires significant manual human effort to create annotations, a very time-consuming and costly process for the map industry. In order to reduce manual annotation burdens, many artificial intelligence (AI) algorithms have been developed to pre-label the HD maps. However, there still exists a large gap between AI algorithms and the traditional manual HD map production pipelines in accuracy and robustness. Furthermore, it is also very resource-costly to build large-scale annotated datasets and advanced machine learning algorithms for AI-based HD map automatic labeling systems. In this paper, we introduce the Tencent HD Map AI (THMA) system, an innovative end-to-end, AI-based, active learning HD map labeling system capable of producing and labeling HD maps with a scale of hundreds of thousands of kilometers. In THMA, we train AI models directly from massive HD map datasets via supervised, self-supervised, and weakly supervised learning to achieve high accuracy and efficiency required by downstream users. THMA has been deployed by the Tencent Map team to provide services to downstream companies and users, serving over 1,000 labeling workers and producing more than 30,000 kilometers of HD map data per day at most. More than 90 percent of the HD map data in Tencent Map is labeled automatically by THMA, accelerating the traditional HD map labeling process by more than ten times.

\end{abstract}

\begin{figure*}[h]
  \centering
  \includegraphics[width=0.9\linewidth]{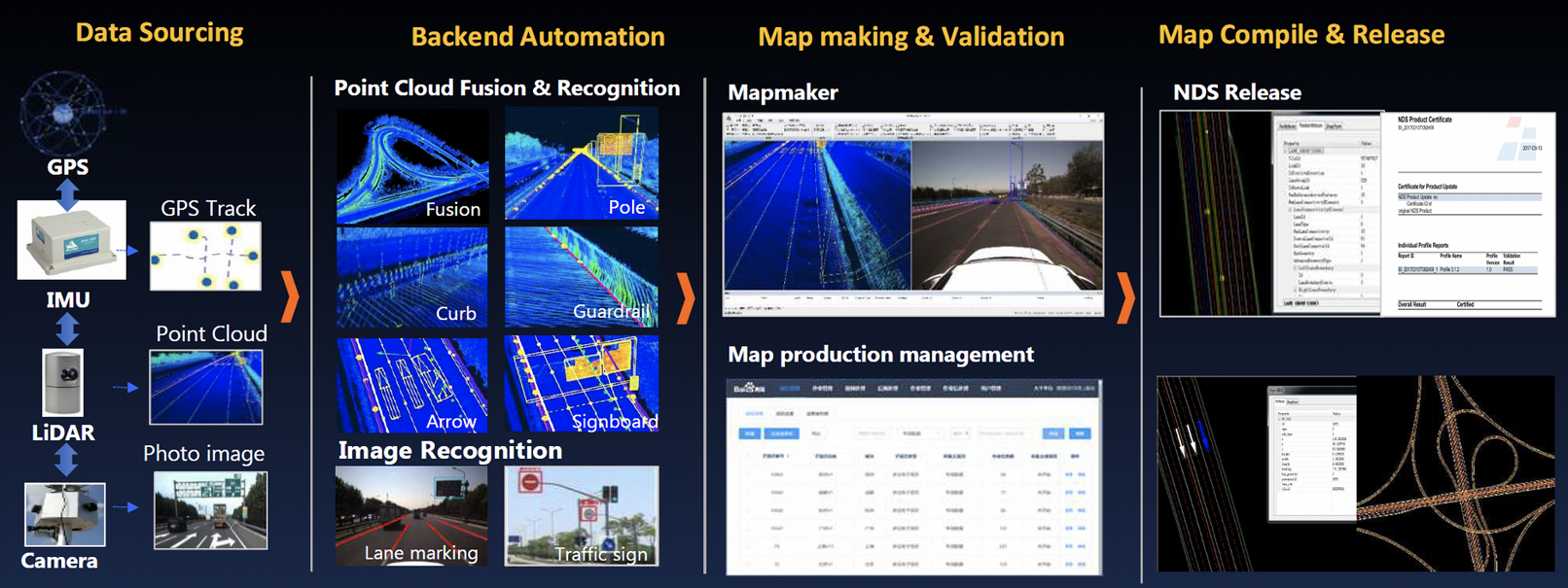} 
  \caption{Basic production process of HD maps: (1) Data sourcing; (2) Backend Automation; (3) Map making and validation; (4) Map compile and release. Example shown uses the Baidu Apollo HD Map production~\citep{baidu_apollo}.}
  \label{fig:architecture_fig1}
\end{figure*}

\section{Introduction}
With the fast development of intelligent transportation, environment perception has become a key factor for autonomous driving. In recent years, various deep neural networks (DNN) have been developed to solve the automatic traffic scene understanding problem, including segmentation-based and object detection-based methods~\citep{fernandes2021point,tang2021review,yan2020lane}. However, developing a robust framework suitable for most road scenarios continues to be extremely challenging since real-world environments exhibit extreme weather variations and obstacles, which significantly affect the final detection results from real-time data. Furthermore, analysis needs to be performed in real time, adding to the challenge. To solve these problems, the current industry standard relies on high definition (HD) maps, a type of centimeter-level imagery collected using a laser sensor, since HD maps contain significantly more detailed representations~\citep{mattyus2016hd,elhousni2020automatic,fan2018baidu,bao2022high} and true ground-absolute accuracy, while are less affected by the driving environment, unlike conventional RGB real-time traffic scene imagery. Specifically, HD maps can provide the user with permanent road elements such as lane marking types within annotated 3D point clouds. Compared with real-time road images, HD maps offer offline centimeter-level location service and significant prior knowledge about traffic scenes for the self-driving vehicle to avoid environmental interference. 

The HD map production process is shown in Figure~\ref{fig:architecture_fig1}. The process consists of four steps: (1) data sourcing, (2) backend automation, (3) map making \& validation, and (4) map compile \& release. The data sourcing is obtained from sensors on the surveying car: Global Positioning System (GPS), Inertial Measurement Unit (IMU), LiDAR, and camera~\citep{bao2022high}. GPS and IMU provide precise absolute localization of tracks. LiDAR, the most crucial sensor for HD maps, collects object location information with centimeter-level precision. The camera is used to provide the RGB image, which is used to detect attributes of the HD map data. The raw point cloud and image data collected from the sensor are fed into the mid-process system, which includes the point cloud fusion and automatic labeling system, consisting of AI and computer vision analysis techniques for both point cloud and images~\citep{elhousni2020automatic,pannen2020keep}. After pre-labeling, the point cloud and pre-labeling data is verified by the HD map maker in the map-making process. In the end, the HD map data is compiled and released. 

The map-making is the most resource-consuming step, and the research community has tried to use DNNs to build automatic AI systems for the labeling process of HD maps~\citep{jiao2018machine,elhousni2020automatic,zhou2021automatic,kim2021hd,li2022hdmapnet}. These methods have achieved relatively good results on simple 2D tasks such as lane marking and road detection. However, the main challenge for existing automated AI solutions is creating HD maps for 2D ground and 3D aerial element annotations in densely populated cities, where maps exhibit noise and many overlapping 3D objects. On the other hand, users of HD maps require accurate maps from these urban environments to generalize. 

In this work, we present the Tencent HD Map AI (THMA) system, an innovative AI-based system for rapidly labelling large collections of HD maps, that has been deployed by the Tencent Map team since 2021 and so far, has served over one thousand users. In particular, Tencent Map smart city applications have used the products, and the automatically labeled HD maps have been provided to downstream self-driving companies. THMA has helped the map makers significantly improve their operational efficiency and reduce HD map annotation costs. To the best of our knowledge, our THMA is one of the industry's most advanced tools for creating HD map annotations, and offers the following advantages:
\begin{itemize}
\item  \emph{Low cost.} Benefitting from self-supervised and weakly supervised pre-training frameworks, THMA forms a closed loop between generating annotations and model training. As a result, THMA can effectively reduce the need for large-scale manual annotation in HD maps.

\item  \emph{End-to-end training pipeline for any HD map scenes.} Compared to the existing HD map automatic labeling system, THMA can create annotations for all sophisticated 2D ground and 3D aerial elements in the next generation HD map system.

\item  \emph{Modular design.} THMA adopts a modular design and consistently meets downstream users' needs, offering a complete, ready-for-integration solution.
\end{itemize}

\section{Overview and Advantages of THMA}
\begin{figure*}[h]
  \centering
  \includegraphics[width=0.9\linewidth]{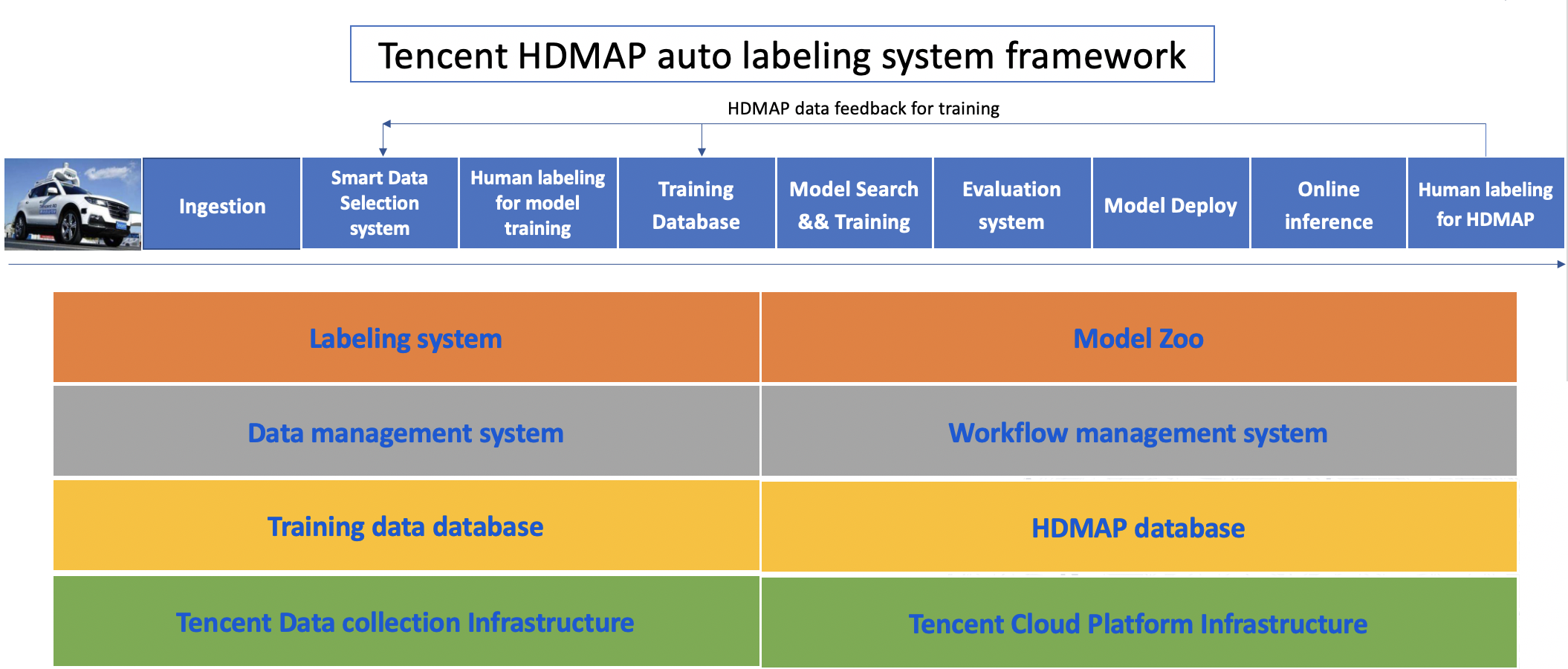} 
  \caption{Detailed overview of the proposed Tencent HD Map AI (THMA) labeling system. The system is modular and is designed to accommodate challenges of labelling large volumes of HD maps of high-density urban environments.}
  \label{fig:labellingSteps_fig2}
\end{figure*}

In this section, we describe the THMA system workflow. THMA was designed for annotating hundreds of thousands of kilometers high-density urban environments, such as China's densely populated cities: Beijing, Shanghai, and Shenzhen (each with a population of greater than $10,000,000$ people), which poses an extremely challenging task. The result is that THMA has a modular workflow, and the key components are shown in Figure~\ref{fig:labellingSteps_fig2}. 

\subsubsection{Next Generation HD Maps}
Specifically, THMA generates annotations for next generation HD maps, described below, delivering highly accurate, up-to-date, and realistic representations of traffic scenery. Next generation HD maps are explicitly built for Level 4 and Level 5 self-driving vehicles, include more abundant and fine-grained traffic scene information than existing deployed HD maps, and will be widely adopted sooner by advanced self-driving systems. Below, we explain key scene attributes (ground elements, separating facilities and 3D aerial elements) of the THMA next generation HD map in detail.

\begin{itemize}
\item \noindent \emph{Ground elements.} Previous work mainly applied old semantics segmentation deep learning algorithms such as FCN~\citep{long2015fully} and U-Net~\citep{ronneberger2015u} to identify a small number of ground elements in HD map systems~\citep{elhousni2020automatic}. We extend the attribute detection of lane markings to 20 types and propose the detection of lane marking attribute change points, road waiting areas, stop lines, and ground traffic signs.

\item \noindent \emph{Separating facilities.} In our work, we add road separating facility modules to detect guardrails, curbs, and natural boundaries. These elements will help improve the safety of automated driving systems.

\item \noindent \emph{3D aerial elements.} The main features and detection difficulties of 3D aerial elements are the distribution diversity of scale and shape. THMA contains more 3D aerial elements. It includes large-scale objects such as tunnels and small-scale objects such as traffic lights. In terms of shape, there are linear objects such as straight poles and curved ones, planar objects such as traffic signs, and thick objects such as traffic lights and tunnels. We adopt a unified end-to-end framework when building THMA. The framework outputs a unified descriptor adapted to the diversity of object shapes and the diversity of the number of objects at the exact location.

\end{itemize}

\begin{figure}[h]
\centering
\subfigure[] 
{
	\centering          
	\includegraphics[width=0.36\linewidth]{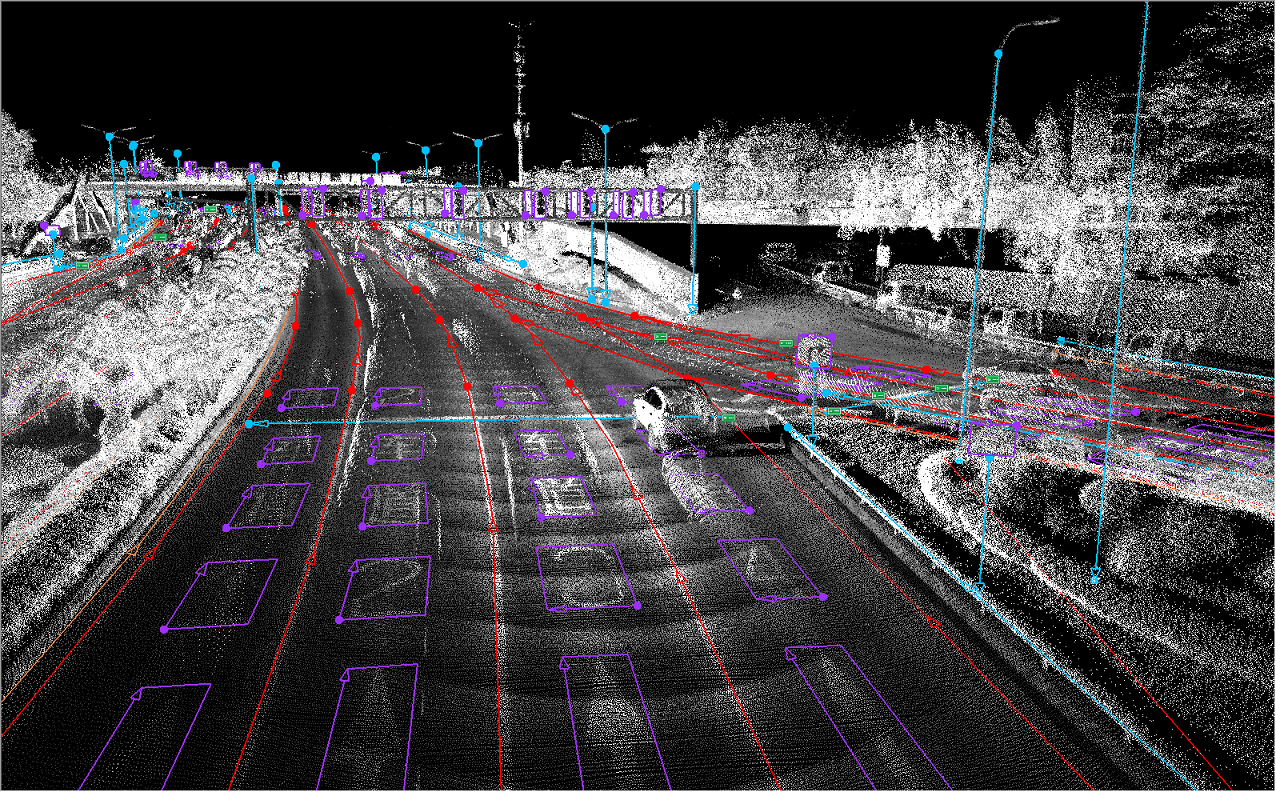}
}
\subfigure[]
{
	\centering     
	\includegraphics[width=0.45\linewidth]{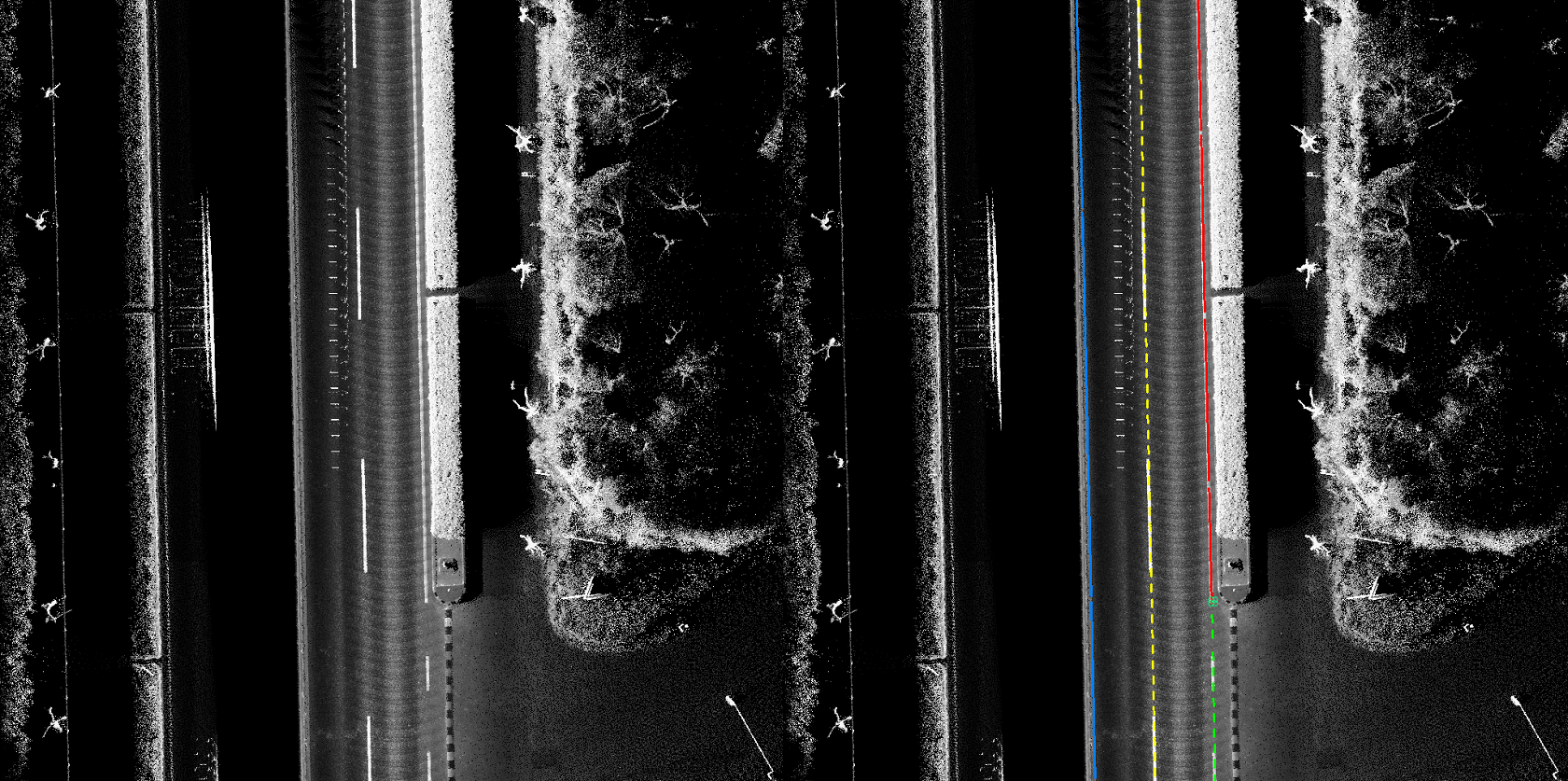}  
}
\caption{(a) Example of manually annotated 3D point cloud. (b) Example of projected Bird's-Eye-View (BEV) images with manually annotated lane markings.}
\label{fig:data_example}
\end{figure}

\begin{figure*}[h]
\centering
\subfigure[] 
{
	\centering          
	\includegraphics[width=0.8\linewidth]{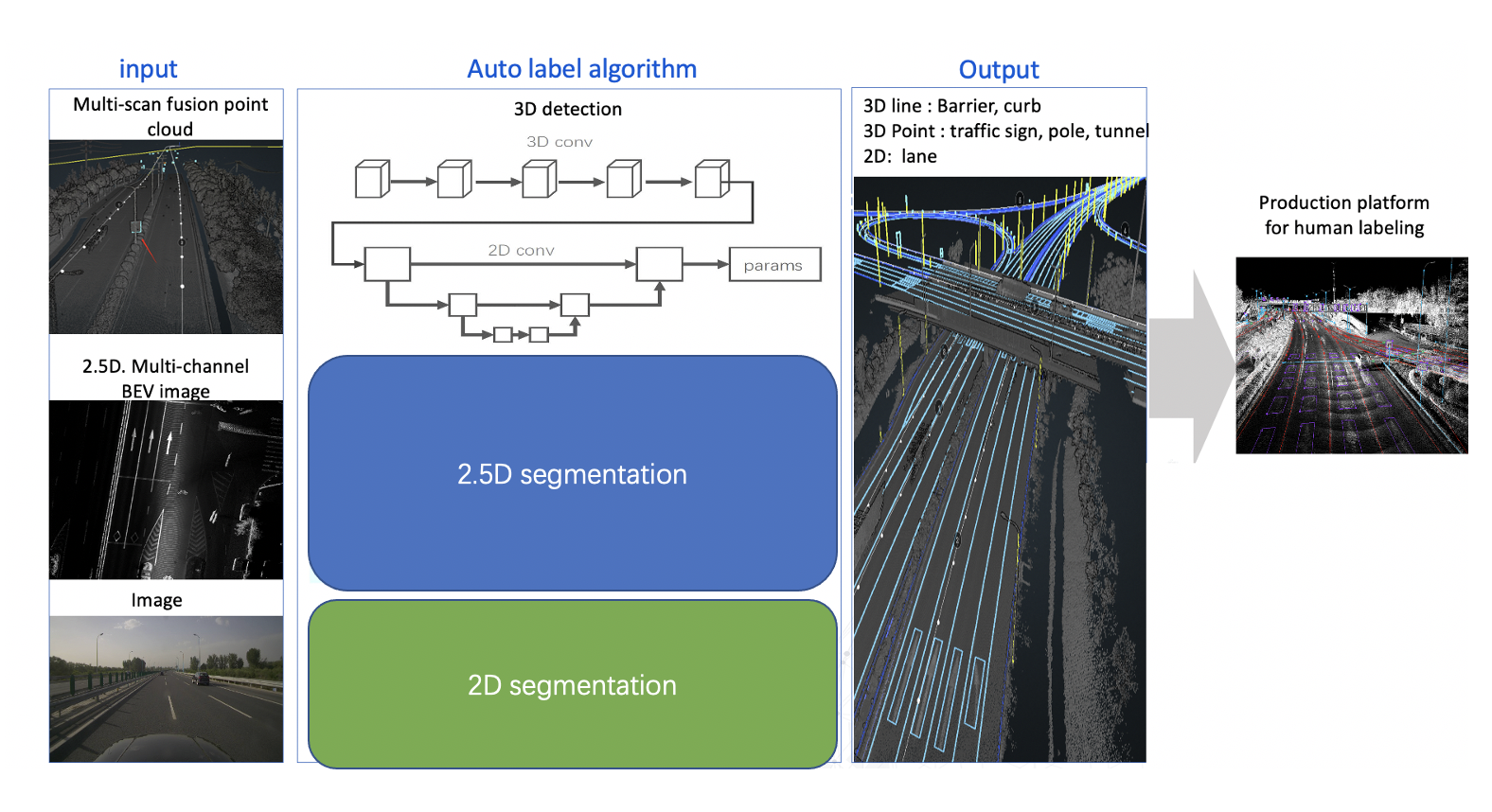}
}
\subfigure[] 
{
	\centering          
	\includegraphics[width=0.45\linewidth]{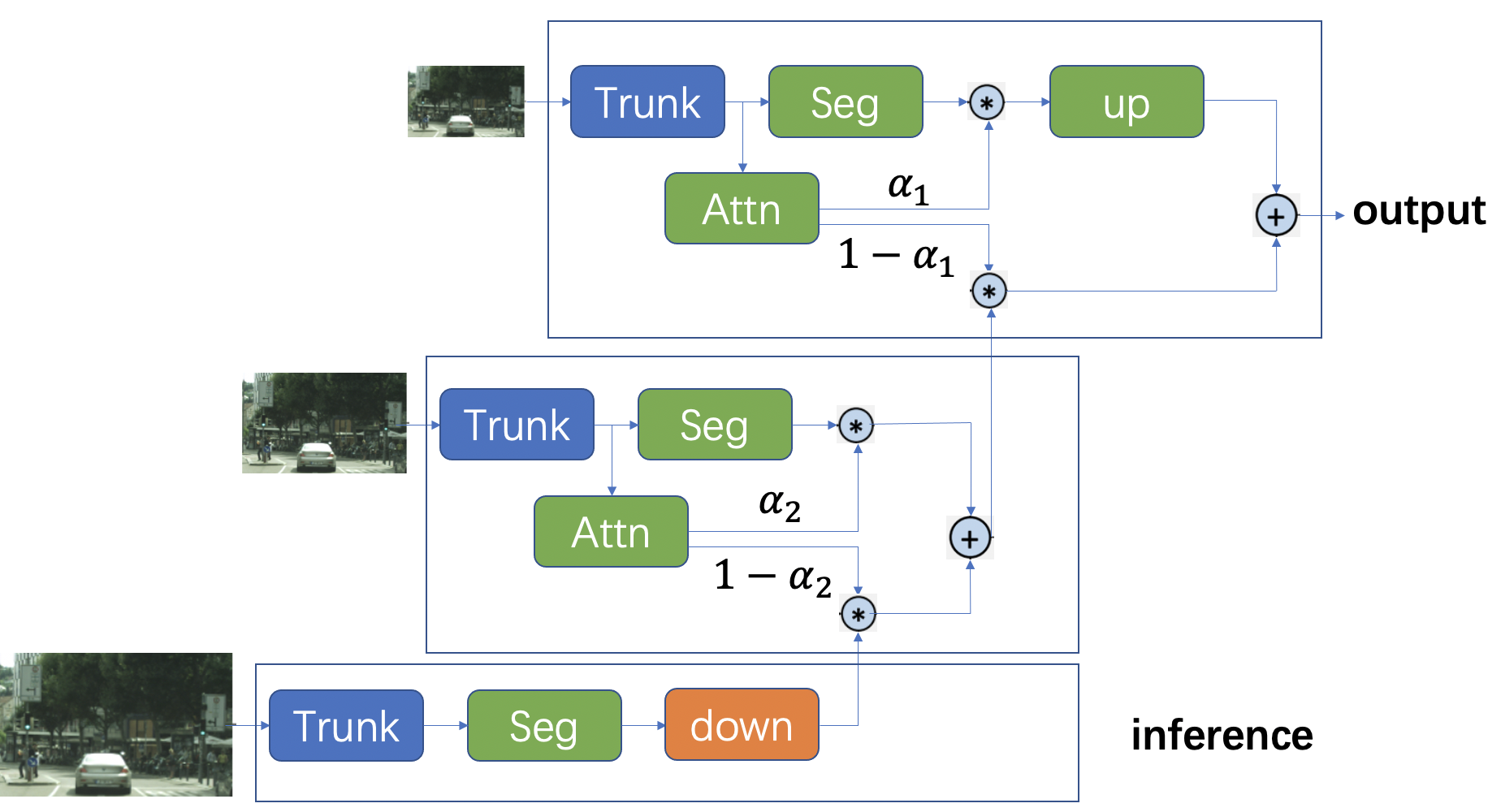}
}
\subfigure[]
{
	\centering     
	\includegraphics[width=0.45\linewidth]{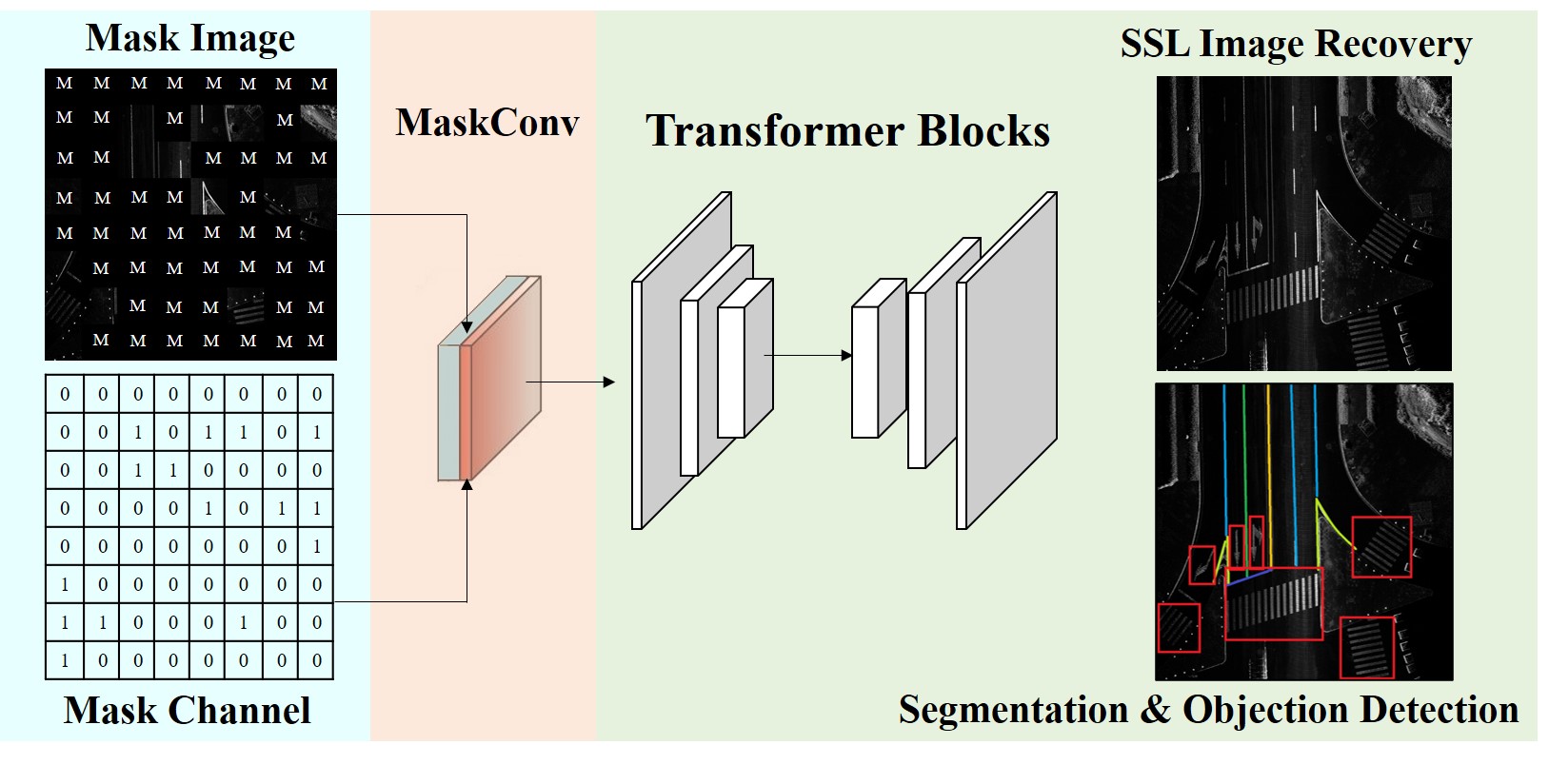}  
}
\caption{(a) Smart data processing in THMA detects objects in 3D point clouds, 2.5D Bird's-Eye-View (BEV) images, and RGB images, fusing resulting aerial and ground object detection results for higher accuracy. (b) 2D segmentation framwork~\citep{tao2020hierarchical}.(c) 2.5D self-supervised learning segmentation framwork.}
\label{fig:smart_framework}
\end{figure*}

\subsubsection{Advantages of AI Driven Annotation}
THMA relies on AI to scale annotations to extremely large volume datasets. The resulting production pipeline is depicted in Figure~\ref{fig:labellingSteps_fig2} and provides not only scalable infrastructure for training and inference but also a centralized data platform for access to meta-data. Once we have the confidence results of AI output, we could save the high-confident output into the HD map production line and send the low-confident output to annotation experts to judge and re-label. The updated labels for low-confident results then go back into the HD maps and are fed back to re-train the AI models in the next iteration. Thus, our AI component can be considered an end-to-end active learning loop~\citep{jiao2018machine,haussmann2020scalable}. The advantages of this active learning system are:
\begin{itemize}
\item \emph{Massive data collection and cloud computation.} Hundreds of thousands of kilometers of HD map raw data is collected through several data-collecting vehicles. In this way, the HD map is updated quickly. To process more data and train new models, Tencent Cloud, a secure, reliable, and high-performance cloud computing service provided by Tencent, is used.

\item \emph{Diverse training database.} Due to the novelty of the framework, the training relies on over 400,000 kilometers of HD map data which contains partially incomplete and inaccurate annotations for self-supervised learning and weakly supervised learning, further enabling model generalization. 

\item \emph{Data and workflow management platform.} With the powerful Tencent Cloud platform, high parallelism, traceability, caching workflow, and PB-level data management have been enabled.

\item \emph{Completed labeling platform.} The labeling platform consists of the HD map label platform, which produces the 3D HD map format data, and the traditional labeled tools which could produce the detection and segmentation data for the 2D or 3D model.  

\item \emph{Powerful model zoo}. The model zoo has up-to-date 2D and 3D detection and segmentation models implemented by PyTorch. In order to make the system end-to-end and generalizable, the model decoders have been redesigned to adapt to the HD map data format. Many self-supervised and weakly supervised methods have been implemented, trained on Tencent Map's GPU clusters and deployed in the cloud.
\end{itemize}

In the next two sections, we describe the data acquisition and the smart data pre-processing (AI-based segmentation and detection steps).  



\section{Data Acquisition}
We first introduce the collection of raw 3D point cloud data and the generation process for and 2.5D Bird's-Eye-View (BEV) images. 
The raw data of the Tencent HD map includes RGB images, GPS position and attitude data, and the laser 3D point cloud. The original 3D point cloud data used in our model training is collected from the newest laser scanner, installed at the tail at a 45-degree angle, focusing on scanning the road surface. Compared with other HD map datasets, our dataset has the advantages of high density, apparent distinction between light and dark reflection intensity, and apparent visual features of ground elements. Besides, all 3D point clouds in our dataset were collected from more complex traffic scenes, such as China's densely populated cities: Beijing, Shanghai, and Shenzhen, each with a population of $>10$ million people. Our traffic scenes include highways, urban expressways, ordinary urban roads, feeder roads, rural roads, tunnels, interchanges, etc. This imagery is currently not well represented in other HD map systems such as Nuscenes~\citep{caesar2020nuscenes}, Waymo~\citep{sun2020scalability}, Argoverse~\citep{wilson2021argoverse}. Our point cloud scanning focuses on depicting road features with high density, high-resolution and significant visual features of reflection intensity, emphasizing the refined detection of traffic attributes under the HD map production requirements. As a result, THMA generates data which is representative of diverse traffic conditions and is the best source for next generation HD maps.

For detection of 3D aerial elements, the best solution is to analyze (segment and detect objects) in 3D point cloud data. However, for the detection of ground elements, 2.5D BEV images, i.e., top-down parallel projection of 3D cloud points, yield better accuracies and inference speeds. One of the key innovations of THMA is that we can efficiently bind 2.5D BEV images and 3D point clouds. Sample data (3D point clouds and 2.5D BEV images) is shown in Figure~\ref{fig:data_example}.

\subsubsection{2.5D BEV Image Generation} The 2.5D BEV projection images we use are generated from the 3D laser point cloud by a top-down parallel projection with minor modifications, such as car removal. For the original 3D point cloud data, we select the resolution of 0.05 meter and calculate the center coordinates, image range, and point cloud range of each projection image according to the trajectory to determine the coordinate conversion parameters. Next, we convert the point cloud within the selected range to the Mercator coordinate system and perform elevation filtering on the 3D point cloud, to only reserve the points near the ground. For the points falling within each pixel, we put the reflection intensity value, the highest elevation value, and the lowest elevation values into the three channels of the 2.5D BEV output, respectively, and normalize the pixel range to 0-255. 

The 2.5D BEV projection images generated through the above process contain rich texture information. Each image is rotated in the direction of vehicle travel. The resulting image size is 1024 $\times$ 1024 and the pixel resolution is 0.05 meters. Considering the quality and gray scale enhancement of the original point cloud, the reflection intensity channel of the BEV image can clearly and better reflect texture characteristics of the road surface. Semantic information such as lane markings, ground signs, and zebra crossings in traffic scenes can be distinguished according to the light and dark changes of the reflection intensity. Besides, each pixel records the highest and the lowest elevation values, respectively, to differentiate the ground and curbs and guardrails, which are challenging to detect from the single channel 2D BEV images.

\section{Smart Data Processing}
The next step in the THMA pipeline, shown in Figure~\ref{fig:smart_framework}, is known as the divide-and-conquer smart data processing, which identifies objects in 3D multi-scan fusion point clouds, 2.5D BEV images, and RGB images. The 3D detection algorithm auto labels 3D points (for traffic lights, poles, tunnels, and traffic signs) and lines (for barriers and curbs) from the multi-scan fusion point cloud. The 2.5D segmentation algorithm detects ground elements such as lane marking on the multi-channel BEV image. The 2D segmentation algorithm detects other attributes, such as the lane marking color in the RGB image. All generated annotations are merged into the final HD map product. Below, we briefly introduce the technical algorithm design for 3D point cloud and BEV elements in HD map automatic annotation.

\subsection{3D Point Cloud Object Detection}
3D objects vary widely in shapes and sizes. Generally, the algorithms on 2D and 3D object detection are based on the detection of the bounding box. However, these algorithms are only suitable for objects with known orientation and aspect ratios. For objects without defined directions, it is difficult or not feasible to define the corners and size of the bounding box. Even if the label is forcibly defined, conflicts between different training samples arise, resulting in non-convergence of training or degradation of the performance of the algorithm. In our case, a unified framework compatible with the diversity of object shapes, sizes and distribution is needed.

Based on the above considerations and previous work in HD map labelling~\citep{yang2018hdnet,yang2018pixor}, we propose a new unified end-to-end 3D model. The schematic diagram is shown in Figure~\ref{fig:smart_framework} 3D point cloud branch and  Figure~\ref{fig:3d_backbone_distillation}. The backbones of the model include 2D and 3D convolutions. The output is a universal descriptor that provides information on the detected objects, instead of just the bounding boxes. In case the object direction cannot be identified, the output descriptor can be used for a unique description without ambiguity. For example, the descriptor does not explicitly define the yaw of a pole. Instead, the apex and bottom points of the pole are provided, and the yaw can be calculated from these quantities. Another example is the traffic cone, which is described using the vertex, center, and radius of the bottom. The corners of the traffic sign can again be computed, according to the previous logic. Finally, the object detected is not required to be thick, flat, rectangular, or even planar.

The resulting model framework is compatible with the diversity of object shapes. Further, we can detect multiple objects (multi-objects) appearing at the same location. Without loss of generality, the output descriptor for multi-objects can be expressed as:

\begin{align}
D_m = s_{0}s_{1}...s_{N}\Vec{V}_{0}\Vec{V}_{1}...\Vec{V}_{N},
\end{align}
where $s_i$ is the activation probability of the corresponding description vector $V_i$, and $V_i$ represents the descriptor for a single object.

\subsubsection{Knowledge Distillation} 
\begin{figure}[h]
\includegraphics[width=0.85\linewidth]{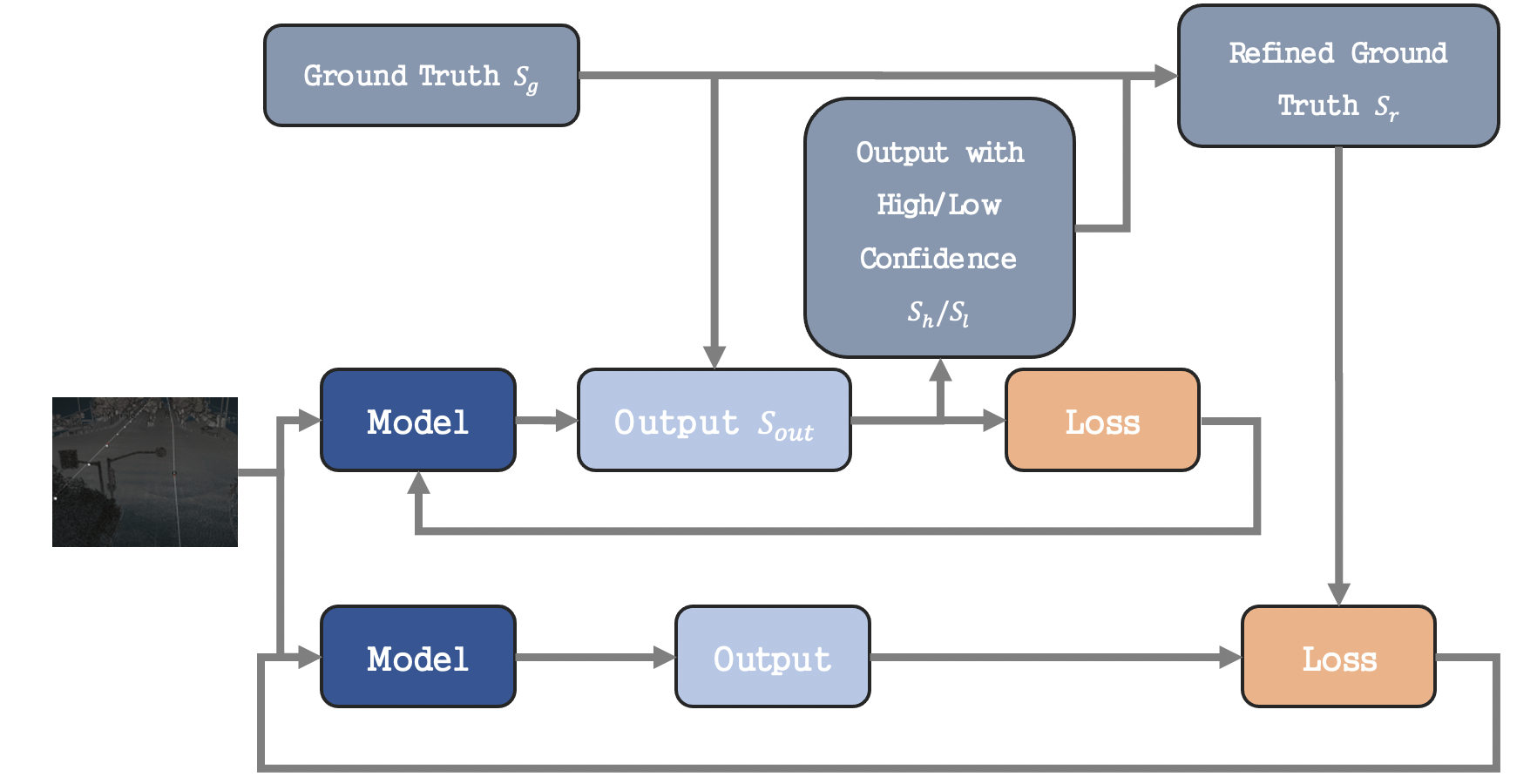}  
\caption{Our solution with knowledge distillation for 3D point cloud object detection.}
\label{fig:3d_backbone_distillation}
\end{figure}

Object labels in 3D cloud points often contain significant noise and labelling errors. These confounding factors influence performance, especially when using focal loss to solve the class imbalance problem during training. To address this challenge, we adopt knowledge distillation in our 3D object detection framework. Knowledge distillation has been proven to yield significant performance improvement for complex 3D point cloud object detection and segmentation tasks~\citep{hou2022point}. Specifically, we construct two training paths. The upper path, shown in the figure~\ref{fig:3d_backbone_distillation}, is the basic model for 3D object detection, including point feature extraction module, point-to-voxel transformation module, encoder-decoder model, etc. The refined ground truth generated by the basic model is combined with the original ground truth and then used as the supervision target for the lower training path. We adopt the output confidence of a positive sample to calculate the difference.

Let $S_g$ be the set of ground truth, e.g., the ground truth bounding boxes, and $S_{out}$ the output of the deep model. The refined ground truth $S_r$ can be computed as:

\begin{align}
    S_r = (S_g \cap S_l) \cup S_h
\end{align}

\begin{align}
    S_l = \{ x | x \in S_{out}, Confidence(x) > T_{low} \}
\end{align}

\begin{align}
    S_h = \{ x | x \in S_{out}, Confidence(x) > T_{high} \},
\end{align}
where $S_l$ is the low confidence result and $S_h$ is the high confidence result.

\subsubsection{Visualization Results} 
\begin{figure}[h]
\centering
\subfigure[] 
{
	\centering          
	\includegraphics[width=0.45\linewidth]{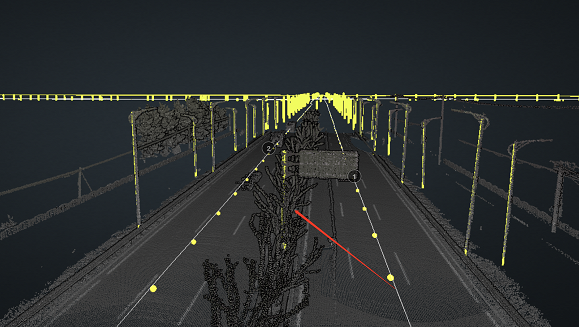}
}
\subfigure[]
{
	\centering     
	\includegraphics[width=0.45\linewidth]{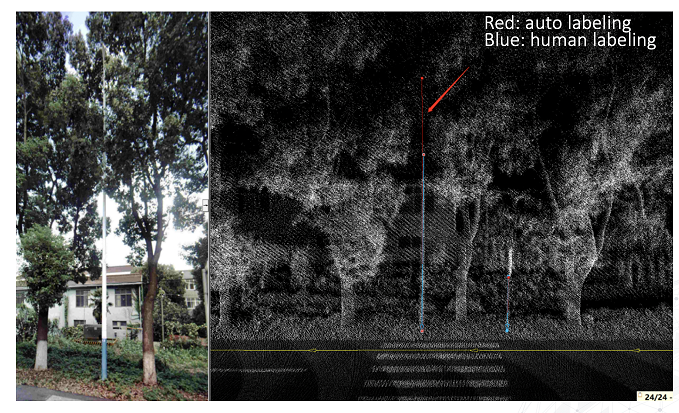} 
}
\caption{Sample label outputs of our system: (a) pole detection results (in yellow), (b) difficult example of pole detection: auto labeling exceeds human labeling ability, labelling (red) part of the pole incorrectly not annotated by a human annotator.}
\label{fig:3D_sample}
\end{figure}

Sample 3D detection results from production are shown in the Figure~\ref{fig:3D_sample}(a) and Figure~\ref{fig:3D_sample}(b). In Figure~\ref{fig:3D_sample}(a), the red arrow indicates the results of our algorithm, which show the challenging example where the pole could be detected correctly even though it is between trees. In some circumstances, the auto label algorithm even performs better than the human annotator, see Figure~\ref{fig:3D_sample}(b). In this example, a part of the pole is occluded by the trees, and the human annotator labeled only the visible part. However, our algorithm correctly labels the missing right top point. The above results demonstrate that the 3D object detection algorithm in THMA is robust and accurate. 

\begin{figure}[h]
\centering
\subfigure[] 
{
	\centering          
	\includegraphics[width=0.45\linewidth]{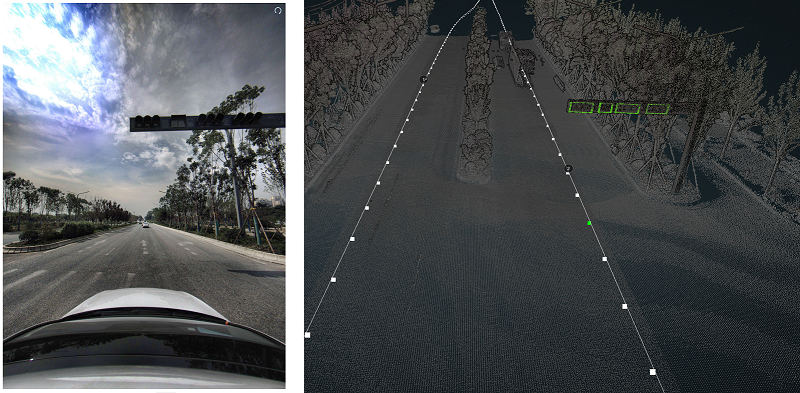}
}
\subfigure[]
{
	\centering     
	\includegraphics[width=0.45\linewidth]{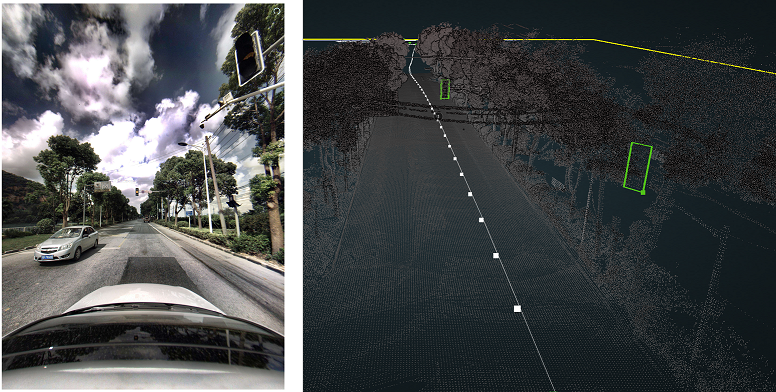}  
}
\caption{(a) Multi adjacent traffic-lights detection results  (b) Diverse angle distribution traffic-lights detection results.}
\label{fig:3D_trafficlight}
\end{figure}

\begin{figure}[h]
 \centering
 \includegraphics[width=0.75\linewidth]{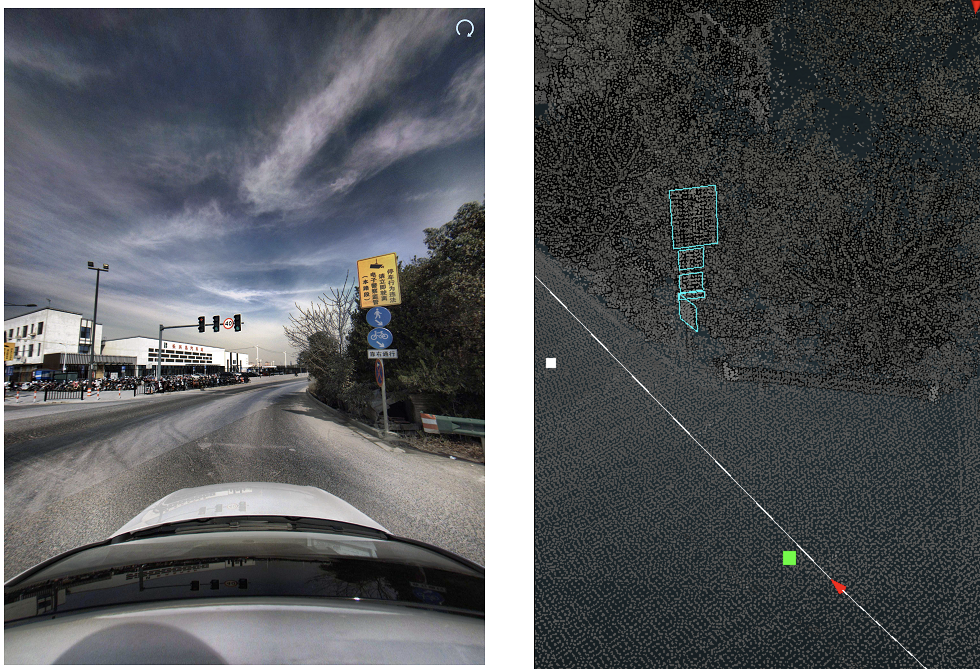} 
 \caption{Sample traffic sign detection results. Our system addresses challenging detection scenarios, such as closely positioned signs.}
 \label{fig:sign_det}
\end{figure}

\begin{figure}[h]
  \centering
  \includegraphics[width=0.75\linewidth]{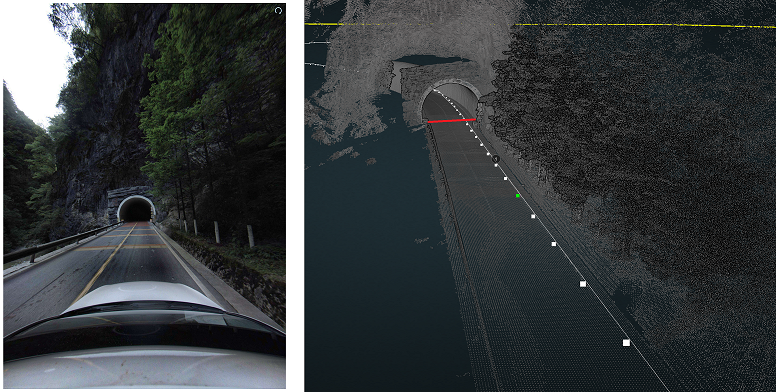} 
  \caption{Sample tunnel detection results. Our system is robust to extreme view-point variations.}
  \label{fig:3D_tunnel}
\end{figure}

We also present additional detection results of traffic lights in Figure~\ref{fig:3D_trafficlight}. The 3D algorithm detects the traffic lights correctly, although they are small and sometimes densely arranged. In Figure~\ref{fig:3D_tunnel} the detection result for the tunnel is shown. Unlike traffic lights, tunnels spread widely in space and a large receptive field is required for their detection. Finally, Figure~\ref{fig:sign_det} shows that our model maintains good results even in the concentrated and complicated traffic sign scenes.

\subsection{2.5D BEV Segmentation and Object Detection}
As discussed above, 2.5D BEV images offer more insightful information for ground object detection, such as better detection and segmentation of lane markings, ground signs and zebra crossings. Below, we discuss the associated analysis steps for 2.5D BEV images.

\subsubsection{Self-Supervised Pre-training}
To address missing labels and noise commonly associated with BEV data, we introduce the newest self-supervised learning methods into our framework. Self-supervised learning aims to design auxiliary tasks that help the model learn meaningful representations from large-scale unlabeled data. We apply the Masked Autoencoder (MAE) technique~\citep{he2022masked}, a popular self-supervised pre-training method, to pre-train the Vision Transformer used in BEV image analysis tasks. The basic structure of MAE is a deep encoder and a lightweight decoder. Only unmasked patches are fed into the encoder, and the decoder processes learn-able masked tokens for image in-painting. Similar to other 3D space and video representation learning tasks~\citep{bao2021beit,feichtenhofer2022masked,tong2022videomae}, we found MAE to be time-saving and effective. 

\begin{figure}[h]
\centering
\includegraphics[width=0.8\linewidth]{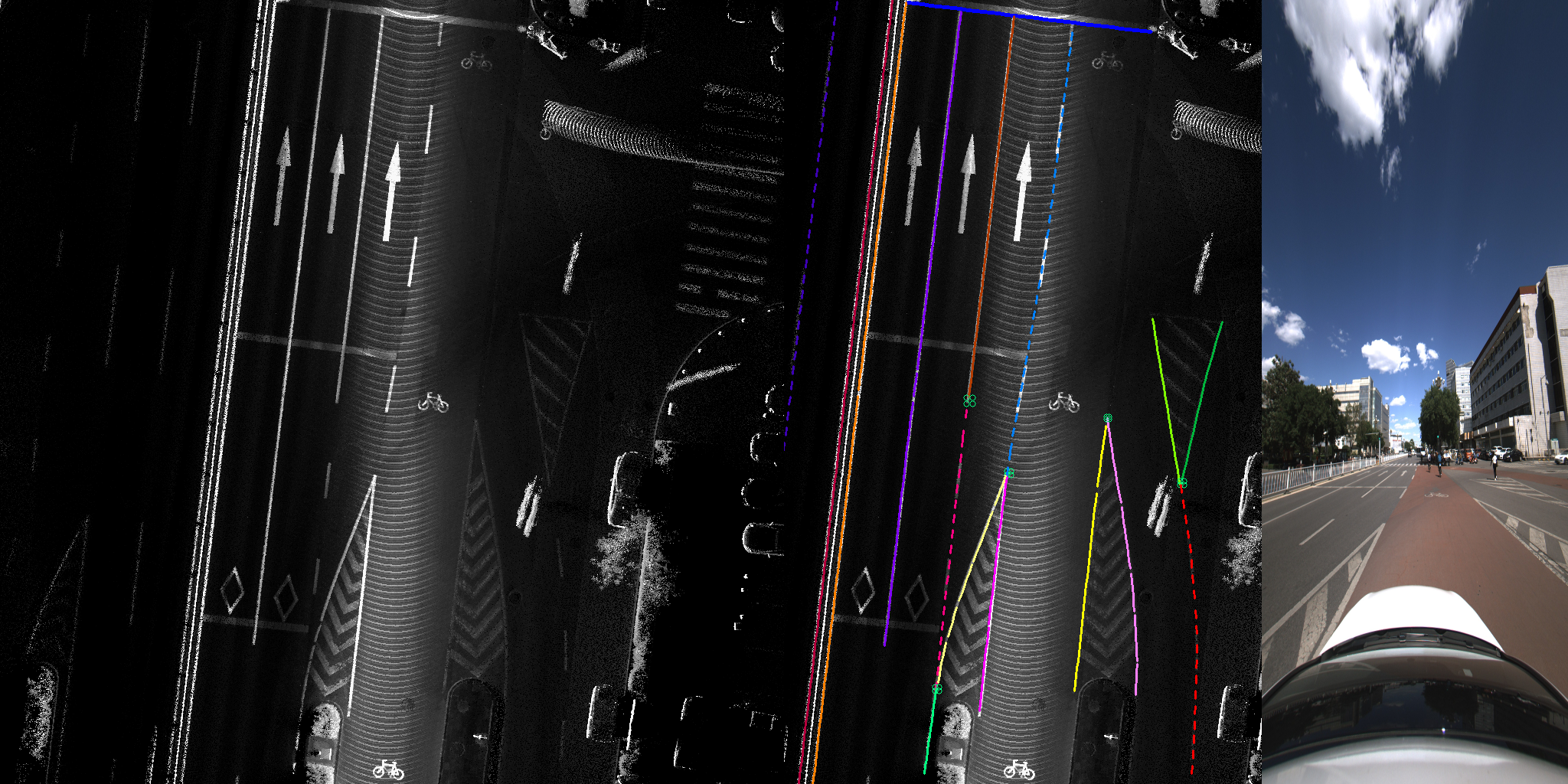}
\caption{Sample results of lane marking detection in BEV imagery from urban areas.}
\label{fig:bev_lane}
\end{figure}

\begin{figure*}[h]
  \centering
  \includegraphics[width=0.9\linewidth]{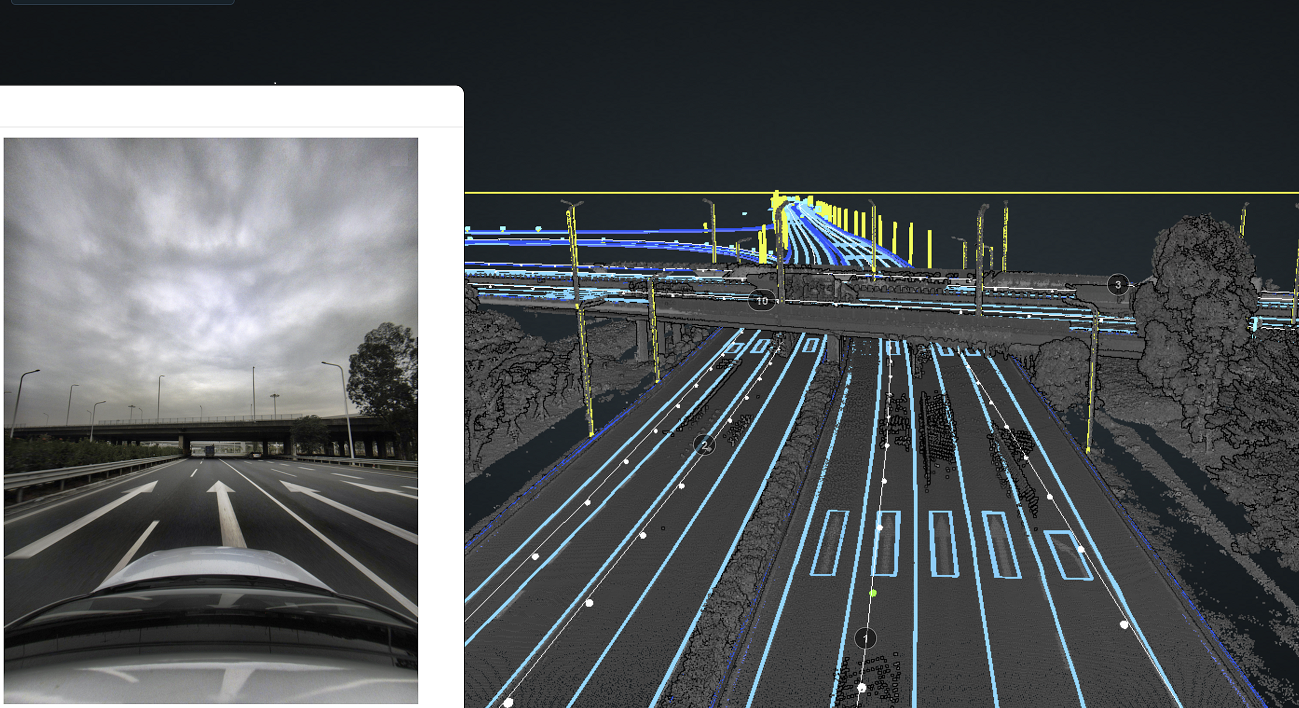} 
  \caption{Qualitative results for the deployed THMA system: the annotations generated from 3D object detection and 2.5D / 2D segmentation and object detection are merged into the HD map system and published to downstream map makers for use.}
  \label{fig:architecture}
\end{figure*}

\subsubsection{Weakly Supervised Pre-training}
In THMA, we are able to leverage large volumes of existing HD maps with corresponding human-annotated results. The challenge in learning AI systems from these datasets is that the human-annotated results have large label noise. In the manual production process, annotation experts can integrate the data according to the production operation specifications and job anti-counterfeiting based on multiple-times data collections. However, for our automatic AI system, the BEV images are generated according to the trajectory 3D point cloud collected at a single time. If we use them directly to generate training data, the model performance will be inadequate due to the label noise problem~\citep{zhou2018brief}. To address this limitation, we generate a large number of incomplete and inaccurate training samples through data mining, non-last night area filtering, and high-reliability area discrimination of the true value of the BEV images. Using these large-scale training samples and limited annotations to pre-train and fine-tine a model on the finely labeled sample set, the robustness in highly complex urban scenes is greatly improved.

\subsubsection{Automatic Ground Element Detection} The final step in analysis of 2.5D BEV images is detection of ground elements, such as lane markings. For all 2D elements, we selected SegFormer~\citep{xie2021segformer}, a segmentation-based vision transformer structure, as the backbone for 2.5D BEV image detection. The key advantage of a Transformer-based method is that the attention map of vision transformer encoders has larger receptive fields than traditional CNN encoders. Different from the original Vision Transformer (ViT)~\citep{dosovitskiy2020image}, SegFormer uses the sequence reduction process to reduce the amount of calculation and accelerates the convergence process during model training~\citep{xie2021segformer, wang2021pyramid}. For each Transformer block, the sequence reduction self-attention is calculated as:

\begin{align}
    Attention(Q, K, V) = Softmax \left (\frac{QK^{T}}{\sqrt{d_{head}}}\right)V
\end{align}

where $K$ is the token representation with initial shape $N \times C$, defined as:

\begin{align}
    K = Linear(\gamma C, C)\cdot\left [ K.Reshape\left (\frac{N}{\gamma}, \gamma C\right) \right ]
\end{align}

The reduction ratio $\gamma$ decreases the dimension of K from $N \times C$ to $N/\gamma \times C$. SegFormer also introduces the $3 \times 3$ depth-wise convolution into the feed-forward network (FFN) to expand the receptive field and reduce harmful effects caused by positional embedding. This is the feed-forward network layer after each self-attention block in SegFormer:

\begin{align}
    y = MLP(Activation[DWConv(MLP(x))]) + x
\end{align}

where $DWConv$ is a $3 \times 3$ depth-wise convolution.

The overall structure of SegFormer consists of a hierarchical transformer encoder and a lightweight MLP decoder which can take advantage of the transformer-induced feature that produces both highly local and non-local attention to rendering powerful representations.

\subsubsection{Visualization Results}

Sample prediction results of the BEV SegFormer are shown in Figure~\ref{fig:bev_lane}. The scenario shown in Figure~\ref{fig:bev_lane} is very complex, including lane markings type change, lane number change, and stop line detection. We not only need to identify the geometric position of the lane markings accurately but also need to accurately detect the attribute of the lane markings and the position at which the number of lanes changes. Benefiting from the self-supervised learning, weakly supervised pre-training, and Vision Transformer, our multitask model can solve the above tasks well.

\section{Application Development and Deployment}

The THMA framework has been developed by the application research team(T, lab) at the Tencent Map starting in 2020. It takes 1.5 years to build and improve the THMA system. In the beginning, this system is an open loop, which will not only need a lot of labeling workers to annotate the training data manually but also decrease the updating frequency. The regular updating frequency for the open loop system is about one month. When we upgrade the system to the closed loop active learning framework, the updating frequency is improved to 1-2 weeks. Besides, the modular design of THMA is combined with the closed-loop active learning framework. When one model for a selected element in the HD map is published or updated (e.g., lane marking color detection), it will first be tested separately and then added to the production pipeline. Note that we used different models for different elements; thus, the whole architecture can be considered as a multi-task framework. Each task can be tested individually, but there are connections between tasks. For example, when detecting the change points of the lane marking attribute, we must use the lane marking position and attribute information obtained by other model branches.

The programming language used in the deployment is Python. To evaluate the model in each version, we use an independent subset of HD maps that are accurately annotated and reviewed by map makers for validation. This subset includes 1,000 kilometers of HD map 3D point clouds, corresponding 3D aerial elements, and 2D ground element annotations. We update the evaluation results for each released version in the product documentation. For overall system performance, we evaluate the performance in terms of automation ratio, i.e., the percentage of HD map data that could be auto-labeled, throughput, i.e., the output volume, as well as the acceleration of the labeling speed. After comparing the labeling results from THMA and human labeling results, the overall automation ratio is more than 90 percent. This way, the labeling speed is accelerated more than ten times. Due to the compact design of the system, the throughput of the system is more than 30,000 kilometers per day.  

\section{Application Impact and Payoff}
THMA has been developed and deployed for two years, used by thousands of annotation experts. To date, our system has produced over 400,000 kilometers of HD map data. It has a record of serving almost one thousand workers to produce 30,000 kilometers of HD map data per day, which is quite advanced to the best of our knowledge. Over the two years of usage, this system has achieved the following business improvements:
\begin{enumerate}
\item \emph{Efficiency.} In the traditional system of auto-labeling, to efficiently develop a model in massive traffic data scenarios such as China, at least several kilometers of point-cloud data and several ten thousands of images are needed, an annotation effort that would require a whole year. Due to the end-to-end data recycling, intelligent data mining, and weakly supervised and self-supervised techniques, THMA reduces the labelling time required by an order of magnitude.

\item \emph{Model Generalization Ability.} Due to processing and learning from hundreds of thousands of kilometers of challenging HD map data, the labelling system has high precision and recall, as well as generalization ability in challenging cases of urban scenery. As a result, THMA creates a record of serving one thousand makers and producing several tens of thousands kilometers per day.

\item \emph{Iterative and Incremental Development.} New requests from downstream smart city applications and self-driving companies are added as time goes by. Since THMA follows a modular design approach around different sub-tasks, product updates can be performed without affecting the overall AI solution. Since deployment in 2021, we release the latest version regularly to customers and our systems are updated every 2 months, on average.
\end{enumerate}

\section{Conclusion and Future Work}

In this paper, we introduce the Tencent HD map AI (THMA) system, a novel, end-to-end, and fully automatic AI system designed to label hundreds of thousands of kilometers of high definition (HD) maps of densely populated urban environments for autonomous driving applications. The system is designed and deployed in production by the Tencent Map T lab team and their users since 2021, generating 30,000 kilometers of HD map data per day and serving over 1,000 labeling workers. To the best of our knowledge, the resulting system is one of the largest in the world to date. The core algorithm propagates annotations from existing Tencent large-scale HD map datasets to newly acquired data and allows for fully automatic and human-in-the-loop type labelling, saving significant time and cost over existing fully manual annotation techniques.

In future work, we plan to expand the existing system focused on lane detection to auto-labelling more complex label relationships. We also hope to leverage iterative and incremental development to further improve robustness. 


\clearpage

\bibliography{aaai23}

\end{document}